\documentclass[conference]{IEEEtran}
\IEEEoverridecommandlockouts

\usepackage{pbalance}

\IEEEoverridecommandlockouts
\usepackage[dvips]{graphicx}
\usepackage{algorithmic}
\usepackage{algorithm}
\usepackage{url}
\usepackage{multirow}

\usepackage{tikz}

\usepackage{bbding}

\usepackage{makecell}
\usepackage{bm}

\title{Task-driven single-image super-resolution reconstruction of document scans\thanks{This work was supported by the National Science Centre, Poland, under Research Grant 2022/47/B/ST6/03009.}}
\author{
\IEEEauthorblockN{Maciej Zyrek, Michal Kawulok}
\IEEEauthorblockA{0009-0009-4709-2743, 0000-0002-3669-5110\\
Department of Algorithmics and Software, Silesian University of Technology\\
Akademicka 16, 44-100 Gliwice, Poland\\
Email: macizyr350@student.polsl.pl; michal.kawulok@polsl.pl}
}

\begin{document}
\maketitle              

\begin{abstract}
Super-resolution reconstruction is aimed at generating images of high spatial resolution from low-resolution observations. State-of-the-art super-resolution techniques underpinned with deep learning allow for obtaining results of outstanding visual quality, but it is seldom verified whether they constitute a valuable source for specific computer vision applications. In this paper, we investigate the possibility of employing super-resolution as a preprocessing step to improve optical character recognition from document scans. To achieve that, we propose to train deep networks for single-image super-resolution in a task-driven way to make them better adapted for the purpose of text detection. As problems limited to a specific task are heavily ill-posed, we introduce a multi-task loss function that embraces components related with text detection coupled with those guided by image similarity. The obtained results reported in this paper are encouraging and they constitute an important step towards real-world super-resolution of document images.
\end{abstract}

\section{Introduction}
\IEEEPARstart{I}{nsufficient} image spatial resolution is often a bottleneck for computer vision systems that limits the capabilities of image analysis algorithms. In order to address that obstacle, considerable research attention has been paid to developing super-resolution (SR) techniques~\cite{YangZhang2019} aimed at reconstructing high-resolution (HR) images from low-resolution (LR) observations, being either a single image~\cite{ChenHe2022} or multiple images presenting the same scene~\cite{Yue2016}.

Potentially, SR algorithms can be extremely valuable in the cases when acquiring an HR image is subject to a trade-off with the acquisition cost (e.g., in remote sensing~\cite{Tarasiewicz2023TGRS}), speed (e.g., in document scanning~\cite{Wang2020TextZoom}), or other factors. However, the attempts to apply SR algorithms as a preprocessing step prior to fulfilling a proper image analysis task are still rather scarce---commonly, the techniques are trained and validated relying on HR reference images, which are downsampled and degraded to simulate the input LR images. As noted in an excellent review by Chen et al.~\cite{ChenHe2022}, deep networks trained from the simulated data render overoptimistic results and their performance in real-world conditions is much worse, when they are applied to enhancing original, rather than downsampled images. There have been some attempts reported to address this problem relying on the use of real-world data for training~\cite{Cai2019,Martens2019}, but acquiring such data is challenging and costly, and it is not straightforward to exploit the HR references when HR and LR images are captured using different sensors~\cite{Kowaleczko2023}. Another possibility to regularize the training performed from the simulated data is to combine the low-level computer vision task of SR reconstruction with high-level ones like semantic segmentation~\cite{GuoWu2019}, object detection~\cite{Haris2021} and recognition~\cite{YangWu2018}. However, this research direction has not been extensively explored so far.

\subsection{Related Work}

Existing SR techniques can be roughly categorized into single-image (SISR)~\cite{ChenHe2022} and multi-image (MISR)~\cite{Yue2016} ones. The latter also embrace methods specialized for video~\cite{LiuRuan2022} and burst-image SR~\cite{Bhat2021}. While MISR techniques underpinned with information fusion are more successful in recovering the actual HR information, they are also definitely more challenging to apply, as multiple images of the same scene must be acquired and co-registered at subpixel precision. As these restrictions turn out to be impractical in many real-life cases, SISR techniques can be straightforwardly applied and they received much larger research attention. With the advent of deep learning, the field of SISR experienced unprecedented advancements~\cite{WangChen2021} which nowadays allow for generating realistic images even at large magnification ratios of $8\times$ and more~\cite{Abiantun2019}. The first convolutional neural network (CNN) for SR (SRCNN)~\cite{Dong2014} already outperformed the techniques based on sparse coding, despite a relatively simple architecture, which was extended and accelerated to create a faster FSRCNN~\cite{Dong2016b}. The subsequent advancements adopted the achievements in feature representation and nonlinear mapping to modeling the relation between LR and HR images~\cite{HuangLi2021}. The larger models included a {very deep SR} (VDSR) network~\cite{Kim2016}, {deep Laplacian pyramid network} (LapSRN) with progressive upsampling~\cite{Lai2018}, enhanced deep SR network (EDSR)~\cite{LimSon2017}, and SRResNet with residual connections~\cite{Ledig2017} which was used as a generator in a generative adversarial network (GAN) setting. The latest trends in SISR are more focused on reducing the size of the deep models, while preserving the reconstruction quality~\cite{Ayazoglu2021}. Recently, it was demonstrated that SISR can benefit from vision transformers~\cite{LuLi2022} which dynamically adjust the size of the feature maps, thus reducing the model complexity.

\begin{figure*}[!ht]
    \centering
    \includegraphics[width=\textwidth]{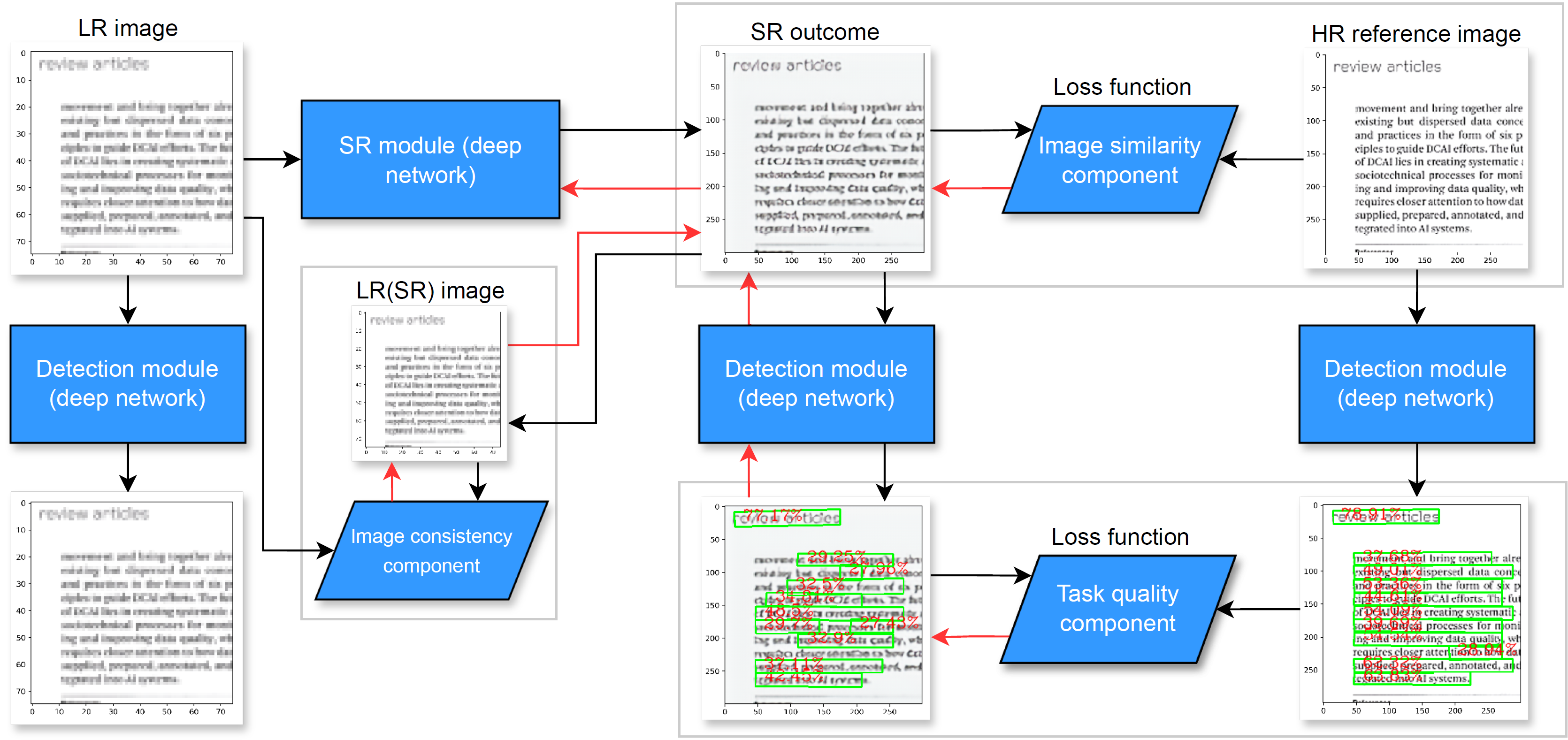}
    \caption{Outline of the proposed self-supervised task-driven training underpinned with text detection. Red arrows indicate the propagation of the loss functions, and the black arrows show the data flow.}
    \label{fig:outline}
\end{figure*}

There have been also some reported attempts to employ SISR to improve text detection and optical character recognition (OCR). Dong et al. adapted their SRCNN for that purpose~\cite{Dong2015} and in~\cite{Pandey2018} a network with a fairly simple architecture with three convolutional layers was employed for super-resolving document scans. Wang et al. proposed to enrich a GAN-based approach with text perceptual loss to help the generator produce recognition-friendly information~\cite{WangXie2019} and later they introduced a TextZoom dataset~\cite{Wang2020TextZoom} composed of cropped texts from the RealSR dataset~\cite{CaiZeng2019RealSR} with natural images captured in uncontrolled environment. In~\cite{ChenLi2021}, a text-focused SR method was introduced which employs a vision transformer to extract sequential information. Inspired by the Gestalt psychology, stroke-based text priors were proposed in~\cite{ChenYu2022} and text priors were exploited for training an SR network in~\cite{MaGuo2023}.

The aforementioned SR techniques were trained to enhance images for OCR, relying on loss functions that are correlated with that specific computer vision task. In addition to that, it is also possible to train an SR network in a task-driven manner, in which the task itself is exploited as a loss function to optimize the network's parameters. Haris et al. applied an {object detection loss}~\cite{Haris2021} which although leads to worse peak signal-to-noise ratio (PSNR) scores than relying on the image-similarity L1 loss, but object detection from the super-resolved images is much more effective. Similar task-driven loss functions were also defined for semantic image segmentation~\cite{Frizza2022,Rad2020}. However, in all these cases the ground-truth references related with the specific task are required for the training data.

\subsection{Contribution}

In this paper, we report our work on task-driven SISR aimed at improving text detection for an OCR system. Our contribution can be summarized as follows:
\begin{enumerate}
    \item We propose a multi-task training underpinned with a loss function composed of image-similarity and text detection-based components that are dynamically balanced throughout the network training.
    \item We propose a self-supervised approach to task-driven training, with the reference labels automatically extracted from the HR reference images.
    \item We report the results of an extensive experimental study which demonstrates that the proposed technique enhances text detection accuracy from document scans super-resolved using three different SR methods.
\end{enumerate}

\section{Proposed Approach}
The proposed task-driven training scheme is outlined in Fig.~\ref{fig:outline}. An SR network is trained using three types of loss functions: (\textit{i})~similarity with the HR reference image, (\textit{ii})~consistency component---similarity between the downsampled SR outcome and the input LR image, and (\textit{iii})~task quality components---the similarity in the space of deep features extracted using a network that performs text detection and recognition. For image-based loss components, we employ the L2 metric (they are termed L2-HR and L2-LR for reference-based and consistency components, respectively), while for computing the task-based loss, we rely on the L1 distance.

In our study, we employed\footnote{For CTPN, we use implementation available at \url{https://github.com/courao/ocr.pytorch}} the connectionist text proposal network (CTPN) for text detection~\cite{TianHuang2016}. 
For CTPN, we exploit 512 features from the final fully-connected layer (we term them as CTPN-deep), as well as the final outputs that encode the coordinates and confidence scores (20 features each, termed CTPN-out). During training, we compute the distances for these three feature spaces and we treat them as different tasks in our setup. For training SR networks, we used a CTPN model that has been already trained---its parameters are frozen during task-driven training and the gradient is propagated to optimize the SR network. Importantly, we establish the target text positions based on the outcome of text detection in the HR reference images. In this way, we do not need the text positions to be annotated, making the training self-supervised.

Our initial attempts to exploit a loss function constructed from multiple components revealed that it is quite challenging to ensure the stability between them during training. Even if we weigh these components to provide a proper initial balance, the training is becoming focused on those that are easier to be optimized and the problem turns into an imbalanced one over time. In order to address that issue, we employed a dynamic weight averaging (DWA) algorithm that adjusts the weights assigned to the particular tasks based on their individual improvements observed in subsequent training steps~\cite{Vandenhende2021}. In DWA, for $N$ tasks, the weight assigned to an $x$-th task at $t$-th step is determined as:
\begin{equation}
    w_x(t) = \left.{N \exp\frac{r_x(t-1)}{T}}\middle/{\sum_{i=1}^{N} \exp\frac{r_i(t-1)}{T}}\right. \rm,
\end{equation}
where
\begin{equation}
    r_i(t)=L_i(t) / L_i(t-1) \rm.
\end{equation}
$L_i$ is the value of the $i$-th loss component and $T$ is the temperature controlling the softness of the task weighting. In this way, the larger weights are assigned to these tasks whose losses have decreases less in the preceding steps.

\begin{table*}[]

    \caption{Quantitative scores obtained for the images from the Old Books and LRDE-DBD benchmarks and from our dataset with document scans, obtained using different SR techniques trained with a variety of loss functions. For each metric and category, the best result is boldfaced.}
    \centering

\renewcommand{\tabcolsep}{1.5mm}
\resizebox{\textwidth}{!}{
    \begin{tabular}{llccccccccccccc}
\Xhline{2\arrayrulewidth}

& & & \multicolumn{4}{c}{\textbf{Loss function}} & & \multicolumn{3}{c}{\textbf{Image similarity metrics}} & & \multicolumn{3}{c}{\textbf{Text detection metrics} } \\ \cline{4-7} \cline{9-11} \cline {13-15}


\multicolumn{3}{l}{\textbf{Model and training type}}  & \textbf{L2-HR} & \textbf{L2-LR} & \textbf{CTPN-deep} & \textbf{CTPN-out} & & \textbf{PSNR}$\uparrow$ & \textbf{SSIM}$\uparrow$ & \textbf{LPIPS}$\downarrow$ & & \textbf{IoU} $\uparrow$ & \textbf{CTPN-deep}$\downarrow$ & \textbf{CTPN-out}$\downarrow$ \\


\multicolumn{3}{r}{(a reference in Fig.~2)} &  & &  &  & & (dB) &  & & &  & ($\cdot 10^{-2}$) & ($\cdot 10^{-2}$) \\



\Xhline{2\arrayrulewidth}



\multicolumn{8}{l}{\textbf{Test set from the simulated benchmark images (Old Books and LRDE-DBD):}} \\

\multicolumn{2}{l}{ SRCNN (from scratch)} & (a) & 	\Checkmark & 	 & 	 & 	 & &	$\bm{21.16}$ & 	$0.8481$ & 	$\bm{0.1818}$ & & 	$0.8923$ & 	--- & 	--- \\ 	
\multicolumn{2}{l}{ SRCNN (from scratch)} & & 	\Checkmark & 	\Checkmark & 	\Checkmark & 	\Checkmark & &	$21.08$ & 	$\bm{0.8489}$ & 	$0.1897$ & & 	$\bm{0.9290}$ & 	$\bm{1.831}$ & 	$\bm{3.366}$ \\ 	\hline
\multicolumn{2}{l}{ FSRCNN (from scratch)} & (b) & 	\Checkmark & 	 & 	 & 	 & &	$24.17$ & 	$\bm{0.9134}$ & 	$\bm{0.1790}$ & & 	$0.9332$ & 	--- & 	--- \\ 	
& --fine-tuned &  &	\Checkmark & 	 & 	\Checkmark & 	\Checkmark & &	$\bm{25.00}$ & 	$0.9071$ & 	$0.2982$ & & 	$\bm{0.9604}$ & 	$1.005$ & 	$1.750$ \\ 	
& --fine-tuned &  &	 & 	\Checkmark & 	\Checkmark & 	\Checkmark & &	$20.06$ & 	$0.6394$ & 	$0.4681$ & & 	$0.9559$ & 	$\bm{0.993}$ & 	$\bm{1.742}$ \\ 	
& --fine-tuned &  &	\Checkmark & 	\Checkmark & 	\Checkmark & 	\Checkmark & &	$24.59$ & 	$0.8848$ & 	$0.3471$ & & 	$0.9560$ & 	$1.113$ & 	$1.939$ \\ 	
\multicolumn{2}{l}{ FSRCNN (from scratch)} & & 	\Checkmark & 	\Checkmark & 	\Checkmark & 	\Checkmark & &	$24.54$ & 	$0.8880$ & 	$0.3245$ & & 	$0.9588$ & 	$1.097$ & 	$1.919$ \\ 	\hline
\multicolumn{2}{l}{ SRResNet (from scratch)} & (c) & 	\Checkmark & 	 & 	 & 	 & &	$24.10$ & 	$0.9147$ & 	$0.1553$ & & 	$0.9392$ & 	--- & 	--- \\ 	
& --fine-tuned &  &	\Checkmark & 	 & 	\Checkmark & 	\Checkmark & &	$28.16$ & 	$0.9537$ & 	$0.1037$ & & 	$0.9614$ & 	$\bm{0.676}$ & 	$\bm{1.177}$ \\ 	
& --fine-tuned &  &	 & 	\Checkmark & 	\Checkmark & 	\Checkmark & &	$24.67$ & 	$0.8404$ & 	$0.3048$ & & 	$0.9676$ & 	$0.694$ & 	$1.198$ \\ 	
& --fine-tuned & (d)  &	\Checkmark & 	\Checkmark & 	\Checkmark & 	\Checkmark & &	$\bm{28.04}$ & 	$\bm{0.9578}$ & 	$\bm{0.0993}$ & & 	$\bm{0.9761}$ & 	$0.714$ & 	$1.235$ \\ 	
\multicolumn{2}{l}{ SRResNet (from scratch)} & & 	\Checkmark & 	\Checkmark & 	\Checkmark & 	\Checkmark & &	$25.49$ & 	$0.9302$ & 	$0.1614$ & & 	$0.9590$ & 	$1.036$ & 	$1.802$ \\ 	
\multicolumn{2}{l}{ SRResNet (from scratch)} & (e) &	 & 	 & 	\Checkmark & 	\Checkmark & &	$2.97$ & 	$-0.1832$ & 	$0.7714$ & & 	$0.9197$ & 	$2.564$ & 	$4.737$ \\

\Xhline{2\arrayrulewidth}

\multicolumn{5}{l}{\textbf{Scanned documents:}} \\

\multicolumn{2}{l}{ SRCNN (from scratch)} & (a) & 	\Checkmark & 	 & 	 & 	 & &	$16.83$ & 	$0.5709$ & 	$\bm{0.4301}$ & & 	$0.7103$ & 	--- & 	--- \\ 	
\multicolumn{2}{l}{ SRCNN (from scratch)} & & 	\Checkmark & 	\Checkmark & 	\Checkmark & 	\Checkmark & &	$\bm{17.06}$ & 	$\bm{0.5782}$ & 	$0.4344$ & & 	$\bm{0.7275}$ & 	$\bm{2.827}$ & 	$\bm{5.342}$ \\ 	\hline
\multicolumn{2}{l}{ FSRCNN (from scratch)} & (b) & 	\Checkmark & 	 & 	 & 	 & &	$18.68$ & 	$\bm{0.6542}$ & 	$0.3681$ & & 	$0.7341$ & 	--- & 	--- \\ 	
& --fine-tuned &  &	\Checkmark & 	 & 	\Checkmark & 	\Checkmark & &	$\bm{18.84}$ & 	$0.6467$ & 	$\bm{0.3585}$ & & 	$0.7608$ & 	$2.363$ & 	$4.290$ \\ 	
& --fine-tuned &  &	 & 	\Checkmark & 	\Checkmark & 	\Checkmark & &	$16.39$ & 	$0.4796$ & 	$0.4343$ & & 	$\bm{0.7688}$ & 	$\bm{2.294}$ & 	$\bm{4.174}$ \\ 	
& --fine-tuned &  &	\Checkmark & 	\Checkmark & 	\Checkmark & 	\Checkmark & &	$18.82$ & 	$0.6429$ & 	$0.3588$ & & 	$0.7635$ & 	$2.328$ & 	$4.219$ \\ 	
\multicolumn{2}{l}{ FSRCNN (from scratch)} & & 	\Checkmark & 	\Checkmark & 	\Checkmark & 	\Checkmark & &	$18.82$ & 	$0.6444$ & 	$0.3644$ & & 	$0.7641$ & 	$2.414$ & 	$4.348$ \\ 	\hline
\multicolumn{2}{l}{ SRResNet (from scratch)} & (c) & 	\Checkmark & 	 & 	 & 	 & &	$18.70$ & 	$0.6634$ & 	$0.3798$ & & 	$0.7264$ & 	--- & 	--- \\ 	
& --fine-tuned &  &	\Checkmark & 	 & 	\Checkmark & 	\Checkmark & &	$19.62$ & 	$0.7075$ & 	$0.3189$ & & 	$\bm{0.7910}$ & 	$\bm{1.985}$ & 	$3.397$ \\ 	
& --fine-tuned &  &	 & 	\Checkmark & 	\Checkmark & 	\Checkmark & &	$19.00$ & 	$0.6327$ & 	$0.3261$ & & 	$0.7886$ & 	$1.994$ & 	$3.520$ \\ 	
& --fine-tuned & (d)  &	\Checkmark & 	\Checkmark & 	\Checkmark & 	\Checkmark & &	$\bm{19.81}$ & 	$\bm{0.7076}$ & 	$\bm{0.3164}$ & & 	$0.7807$ & 	$2.023$ & 	$\bm{3.483}$ \\ 	
\multicolumn{2}{l}{ SRResNet (from scratch)} & & 	\Checkmark & 	\Checkmark & 	\Checkmark & 	\Checkmark & &	$19.23$ & 	$0.6731$ & 	$0.3591$ & & 	$0.7576$ & 	$2.361$ & 	$4.280$ \\ 	
\multicolumn{2}{l}{ SRResNet (from scratch)} & (e) &	 & 	 & 	\Checkmark & 	\Checkmark & &	$2.91$ & 	$-0.0929$ & 	$0.9250$ & & 	$0.7186$ & 	$3.170$ & 	$5.592$ \\

\Xhline{2\arrayrulewidth}

    \end{tabular}
}

    \label{tab:scores}
\end{table*}

\begin{figure*}
    \centering
    \includegraphics[width=\textwidth]{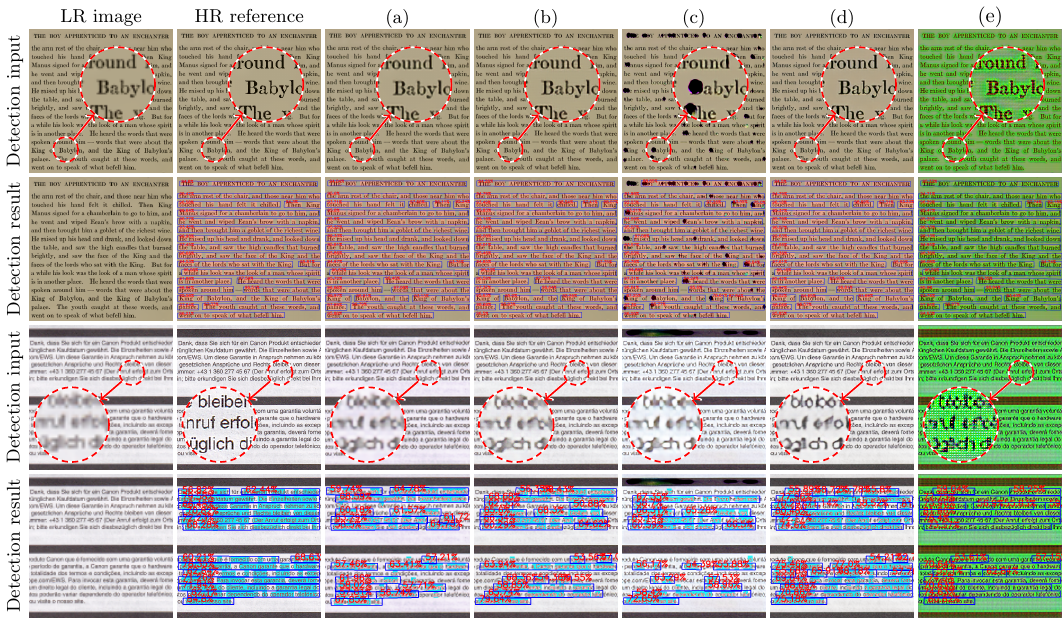}
    \caption{Example of SR reconstruction performed using: (a)~SRCNN, (b)~FSRCNN and (c)~SRResNet models, all trained with the L2-HR loss function, (d)~fine-tuned SRResNet model using all loss functions (L2-HR, L2-LR, CTPN-deep and CTPN-out), and (e)~SRResNet trained from scratch with the task-based CTPN-deep and CTPN-out loss functions. These settings are also referred to in Table~\ref{tab:scores}. We present the examples from the Old Books dataset (two upper rows) and from our dataset with scanned documents (two bottom rows). For each example, we include the detection input (hence the SR outcome) and the result of text detection.}
    \label{fig:example}
\end{figure*}

\section{Experiments}

In our experiments, we exploited three types of datasets: (\textit{i})~natural MS COCO images~\cite{LinMaire2014} for training baseline SR models, (\textit{ii})~scans from the benchmark datasets: Old Books\footnote{Available at \url{https://github.com/PedroBarcha/old-books-dataset}} and LRDE Document Binarization Dataset (LRDE-DBD)\footnote{Available at \url{https://www.lrde.epita.fr/wiki/Olena/DatasetDBD}}~\cite{Lazzara2014}, and (\textit{iii})~our \textit{scanned documents} dataset with real-world scans performed using a Canon LiDE 400 scanner. In our study, we investigated the SRCNN~\cite{Dong2014}, FSRCNN~\cite{Dong2016b} and SRResNet~\cite{Ledig2017} techniques for SR at $4\times$ magnification factor. For training these methods using the regular image-based loss function (L2-HR), we exploited the MS COCO images (LR images were obtained by downsampling the HR images) and for task-based training, we exploited a training set extracted from the Old Books and LRDE-DBD datasets (70\% images). The test sets were formed from the remaining 30\% of Old Books and LRDE-DBD datasets, as well as from all the scanned documents (we used five different scans split into 864 patches). The CTPN model was trained beforehand from the ICDAR2017 dataset~\cite{Gomez2017} and its parameters were frozen during the task-driven trainings.

The reconstruction quality was measured relying on image similarity metrics, namely PSNR, structural similarity index (SSIM), and learned perceptual image patch similarity (LPIPS)~\cite{ZhangIsola2018}, computed between the super-resolved image and the HR reference. For assessing the text detection quality, we employed intersection over union (IoU) between the text positions detected in the super-resolved image and in the corresponding HR reference. We also report the distances in the CTPN-deep and CTPN-out feature spaces that are used for computing the task-based components of the loss function.


First, we trained each of the three networks from scratch (60 epochs), guiding the training using a standard baseline configuration (based on the L2-HR loss) and using all loss components, including L2-HR, the consistency (L2-LR) and task-based CTPN-deep and CTPN-out components. For FSRCNN and SRResNet, we fine-tuned the baseline models (100 epochs) relying on (\textit{i})~L2-HR loss combined with the task-based loss components, (\textit{ii})~the task component coupled with the consistency loss, and (\textit{iii})~using all loss components. In addition to that, we trained SRResNet from scratch relying only on the task-based components (hence without using the image similarity at all). In Table~\ref{tab:scores}, we report the scores obtained for two test sets (unseen during training): for the test set of the benchmark datasets and for our dataset with the scanned documents. It can be observed that incorporating the task-based components improves the scores  in terms of the image-based metrics  in most cases and it always improves the quality of the text detection task (the differences are definitely higher for our scanned documents). It is also clear that the models cannot be trained from scratch without using the image-based components---apparently the problem is not convex enough and the training gets stuck in a local minimum.

A sample of the qualitative results is presented in Fig.~\ref{fig:example} for a benchmark image and one of our scans (the configurations presented in the figure are referenced from Table~\ref{tab:scores}). While for the benchmark image (two upper rows), the text quality is consistently good across all configurations, for our scan, it is definitely better for the model fine-tuned in a task-driven way (d), and it is actually quite close to the result obtained in the HR reference. It can also be seen that the texts are quite clear when SRResNet is trained without using the image-based loss components (which also leads to good detection outcome), but the stability in the color space is not preserved, leading to extremely poor quantitative scores reported earlier in Table~\ref{tab:scores}.

\section{Conclusions and outlook}

In this paper, we reported our initial attempts to apply task-driven training for SISR, guided by text detection. The results are highly encouraging, revealing high potential of task-based loss functions. Importantly, in contrast to the earlier works concerned with task-driven SR, we train the models in a self-supervised way, as we retrieve the annotations by processing the HR reference images.

Our ongoing research is focused on including the text recognition components that may improve the guidance during training. Also, we plan to adapt our approach to MISR problems and to create a dataset embracing multi-image scans.


\begin{thebibliography}{10}
\providecommand{\url}[1]{#1}
\csname url@samestyle\endcsname
\providecommand{\newblock}{\relax}
\providecommand{\bibinfo}[2]{#2}
\providecommand{\BIBentrySTDinterwordspacing}{\spaceskip=0pt\relax}
\providecommand{\BIBentryALTinterwordstretchfactor}{4}
\providecommand{\BIBentryALTinterwordspacing}{\spaceskip=\fontdimen2\font plus
\BIBentryALTinterwordstretchfactor\fontdimen3\font minus \fontdimen4\font\relax}
\providecommand{\BIBforeignlanguage}[2]{{%
\expandafter\ifx\csname l@#1\endcsname\relax
\typeout{** WARNING: IEEEtranS.bst: No hyphenation pattern has been}%
\typeout{** loaded for the language `#1'. Using the pattern for}%
\typeout{** the default language instead.}%
\else
\language=\csname l@#1\endcsname
\fi
#2}}
\providecommand{\BIBdecl}{\relax}
\BIBdecl

\bibitem{Abiantun2019}
R.~Abiantun, F.~Juefei-Xu, U.~Prabhu, and M.~Savvides, ``{SSR2}: Sparse signal recovery for single-image super-resolution on faces with extreme low resolutions,'' \emph{Pattern Recognition}, vol.~90, pp. 308--324, 2019.

\bibitem{Ayazoglu2021}
M.~Ayazoglu, ``Extremely lightweight quantization robust real-time single-image super resolution for mobile devices,'' in \emph{Proc. IEEE/CVF CVPR}, 2021, pp. 2472--2479.

\bibitem{Bhat2021}
G.~Bhat, M.~Danelljan, L.~Van~Gool, and R.~Timofte, ``Deep burst super-resolution,'' in \emph{Proc. IEEE/CVF CVPR}, 2021, pp. 9209--9218.

\bibitem{Cai2019}
J.~Cai, S.~Gu, R.~Timofte, and L.~Zhang, ``{NTIRE 2019 Challenge} on real image super-resolution: {Methods} and results,'' in \emph{Proc. IEEE/CVF CVPR}, 2019, pp. 1--13.

\bibitem{CaiZeng2019RealSR}
J.~Cai, H.~Zeng, H.~Yong, Z.~Cao, and L.~Zhang, ``Toward real-world single image super-resolution: A new benchmark and a new model,'' in \emph{Proc. IEEE ICCV}, 2019.

\bibitem{ChenHe2022}
H.~Chen, X.~He, L.~Qing, Y.~Wu, C.~Ren, R.~E. Sheriff, and C.~Zhu, ``Real-world single image super-resolution: A brief review,'' \emph{Information Fusion}, vol.~79, pp. 124--145, 2022.

\bibitem{ChenLi2021}
J.~Chen, B.~Li, and X.~Xue, ``Scene text telescope: Text-focused scene image super-resolution,'' in \emph{Proc. IEEE/CVF CVPR}, 2021, pp. 12\,026--12\,035.

\bibitem{ChenYu2022}
J.~Chen, H.~Yu, J.~Ma, B.~Li, and X.~Xue, ``Text {Gestalt}: Stroke-aware scene text image super-resolution,'' in \emph{Proc. AAAI Conference on Artificial Intelligence}, vol.~36, no.~1, 2022, pp. 285--293.

\bibitem{Dong2014}
C.~Dong, C.~C. Loy, K.~He, and X.~Tang, ``Learning a deep convolutional network for image super-resolution,'' in \emph{Proc. IEEE/CVF ECCV}.\hskip 1em plus 0.5em minus 0.4em\relax Springer, 2014, pp. 184--199.

\bibitem{Dong2016b}
C.~Dong, C.~C. Loy, and X.~Tang, ``Accelerating the super-resolution convolutional neural network,'' in \emph{Proc. IEEE/CVF ECCV}.\hskip 1em plus 0.5em minus 0.4em\relax Springer, 2016, pp. 391--407.

\bibitem{Dong2015}
C.~Dong, X.~Zhu, Y.~Deng, C.~C. Loy, and Y.~Qiao, ``Boosting optical character recognition: A super-resolution approach,'' \emph{arXiv preprint arXiv:1506.02211}, 2015.

\bibitem{Frizza2022}
T.~Frizza, D.~G. Dansereau, N.~M. Seresht, and M.~Bewley, ``Semantically accurate super-resolution generative adversarial networks,'' \emph{Computer Vision and Image Understanding}, p. 103464, 2022.

\bibitem{Gomez2017}
R.~Gomez, B.~Shi, L.~Gomez, L.~Numann, A.~Veit, J.~Matas, S.~Belongie, and D.~Karatzas, ``ICDAR2017 robust reading challenge on COCO-text,'' in \emph{2017 14th IAPR International Conference on Document Analysis and Recognition (ICDAR)}, vol.~1.\hskip 1em plus 0.5em minus 0.4em\relax IEEE, 2017, pp. 1435--1443.

\bibitem{GuoWu2019}
Z.~Guo, G.~Wu, X.~Song, W.~Yuan, Q.~Chen, H.~Zhang, X.~Shi, M.~Xu, Y.~Xu, R.~Shibasaki \emph{et~al.}, ``Super-resolution integrated building semantic segmentation for multi-source remote sensing imagery,'' \emph{IEEE Access}, vol.~7, pp. 99\,381--99\,397, 2019.

\bibitem{Haris2021}
M.~Haris, G.~Shakhnarovich, and N.~Ukita, ``Task-driven super resolution: Object detection in low-resolution images,'' in \emph{Proc. ICONIP}.\hskip 1em plus 0.5em minus 0.4em\relax Springer, 2021, pp. 387--395.

\bibitem{HuangLi2021}
Y.~Huang, J.~Li, X.~Gao, Y.~Hu, and W.~Lu, ``Interpretable detail-fidelity attention network for single image super-resolution,'' \emph{IEEE Transactions on Image Processing}, vol.~30, pp. 2325--2339, 2021.

\bibitem{Kim2016}
J.~Kim, J.~Kwon~Lee, and K.~Mu~Lee, ``Accurate image super-resolution using very deep convolutional networks,'' in \emph{Proc. IEEE/CVF CVPR}, 2016, pp. 1646--1654.

\bibitem{Kowaleczko2023}
P.~Kowaleczko, T.~Tarasiewicz, M.~Ziaja, D.~Kostrzewa, J.~Nalepa, P.~Rokita, and M.~Kawulok, ``A real-world benchmark for {Sentinel-2} multi-image super-resolution,'' \emph{Scientific Data}, vol.~10, no.~1, p. 644, 2023.

\bibitem{Lai2018}
W.~{Lai}, J.~{Huang}, N.~{Ahuja}, and M.~{Yang}, ``Fast and accurate image super-resolution with deep {Laplacian} pyramid networks,'' \emph{IEEE Trans. on Pattern Analysis and Machine Intelligence}, vol.~41, no.~11, pp. 2599--2613, 2019.

\bibitem{Lazzara2014}
G.~Lazzara and T.~G{\'e}raud, ``Efficient multiscale sauvola's binarization,'' \emph{International Journal on Document Analysis and Recognition (IJDAR)}, vol.~17, no.~2, pp. 105--123, 2014.

\bibitem{Ledig2017}
C.~Ledig, L.~Theis, F.~Husz{\'a}r, J.~Caballero, A.~Cunningham \emph{et~al.}, ``Photo-realistic single image super-resolution using a generative adversarial network,'' in \emph{Proc. IEEE/CVF CVPR}, 2017, pp. 4681--4690.

\bibitem{LimSon2017}
B.~Lim, S.~Son, H.~Kim, S.~Nah, and K.~Mu~Lee, ``Enhanced deep residual networks for single image super-resolution,'' in \emph{Proc. IEEE/CVF CVPR Workshops}, 2017, pp. 136--144.

\bibitem{LinMaire2014}
T.-Y. Lin, M.~Maire, S.~Belongie, J.~Hays, P.~Perona, D.~Ramanan, P.~Doll{\'a}r, and C.~L. Zitnick, ``Microsoft {COCO}: Common objects in context,'' in \emph{Proc. IEEE/CVF ECCV}.\hskip 1em plus 0.5em minus 0.4em\relax Springer, 2014, pp. 740--755.

\bibitem{LiuRuan2022}
H.~Liu, Z.~Ruan, P.~Zhao, C.~Dong, F.~Shang, Y.~Liu, L.~Yang, and R.~Timofte, ``Video super-resolution based on deep learning: a comprehensive survey,'' \emph{Artificial Intelligence Review}, pp. 1--55, 2022.

\bibitem{LuLi2022}
Z.~Lu, J.~Li, H.~Liu, C.~Huang, L.~Zhang, and T.~Zeng, ``Transformer for single image super-resolution,'' in \emph{Proc. IEEE/CVF CVPR}, 2022, pp. 457--466.

\bibitem{MaGuo2023}
J.~Ma, S.~Guo, and L.~Zhang, ``Text prior guided scene text image super-resolution,'' \emph{IEEE Transactions on Image Processing}, vol.~32, pp. 1341--1353, 2023.

\bibitem{Martens2019}
M.~M{\"a}rtens, D.~Izzo, A.~Krzic, and D.~Cox, ``Super-resolution of {PROBA-V} images using convolutional neural networks,'' \emph{Astrodynamics}, vol.~3, no.~4, pp. 387--402, 2019.

\bibitem{Pandey2018}
R.~K. Pandey and A.~Ramakrishnan, ``Efficient document-image super-resolution using convolutional neural network,'' \emph{S{\=a}dhan{\=a}}, vol.~43, pp. 1--6, 2018.

\bibitem{Rad2020}
M.~S. Rad, B.~Bozorgtabar, C.~Musat, U.-V. Marti, M.~Basler, H.~K. Ekenel, and J.-P. Thiran, ``Benefiting from multitask learning to improve single image super-resolution,'' \emph{Neurocomputing}, vol. 398, pp. 304--313, 2020.

\bibitem{Tarasiewicz2023TGRS}
T.~Tarasiewicz, J.~Nalepa, R.~A. Farrugia, G.~Valentino, M.~Chen, J.~A. Briffa, and M.~Kawulok, ``Multitemporal and multispectral data fusion for super-resolution of {Sentinel-2} images,'' \emph{IEEE Transactions on Geoscience and Remote Sensing}, vol.~61, pp. 1--19, 2023.

\bibitem{TianHuang2016}
Z.~Tian, W.~Huang, T.~He, P.~He, and Y.~Qiao, ``Detecting text in natural image with connectionist text proposal network,'' in \emph{Proc. IEEE/CVF ECCV}.\hskip 1em plus 0.5em minus 0.4em\relax Springer, 2016, pp. 56--72.

\bibitem{Vandenhende2021}
S.~Vandenhende, S.~Georgoulis, W.~Van~Gansbeke, M.~Proesmans, D.~Dai, and L.~Van~Gool, ``Multi-task learning for dense prediction tasks: A survey,'' \emph{IEEE Transactions on Pattern Analysis and Machine Intelligence}, vol.~44, no.~7, pp. 3614--3633, 2021.

\bibitem{Wang2020TextZoom}
W.~Wang, E.~Xie, X.~Liu, W.~Wang, D.~Liang, C.~Shen, and X.~Bai, ``Scene text image super-resolution in the wild,'' in \emph{Proc. IEEE/CVF ECCV}.\hskip 1em plus 0.5em minus 0.4em\relax Springer, 2020, pp. 650--666.

\bibitem{WangXie2019}
W.~Wang, E.~Xie, P.~Sun, W.~Wang, L.~Tian, C.~Shen, and P.~Luo, ``{TextSR}: Content-aware text super-resolution guided by recognition,'' \emph{arXiv preprint arXiv:1909.07113}, 2019.

\bibitem{WangChen2021}
Z.~Wang, J.~Chen, and S.~C.~H. Hoi, ``Deep learning for image super-resolution: A survey,'' \emph{IEEE Transactions on Pattern Analysis and Machine Intelligence}, vol.~43, no.~10, pp. 3365--3387, 2021.

\bibitem{YangZhang2019}
W.~{Yang}, X.~{Zhang}, Y.~{Tian}, W.~{Wang}, J.~{Xue}, and Q.~{Liao}, ``Deep learning for single image super-resolution: A brief review,'' \emph{IEEE Transaction on Multimedia}, vol.~21, no.~12, pp. 3106--3121, Dec 2019.

\bibitem{YangWu2018}
X.~Yang, W.~Wu, K.~Liu, P.~W. Kim, A.~K. Sangaiah, and G.~Jeon, ``Long-distance object recognition with image super resolution: A comparative study,'' \emph{IEEE Access}, vol.~6, pp. 13\,429--13\,438, 2018.

\bibitem{Yue2016}
L.~Yue, H.~Shen, J.~Li, Q.~Yuan, H.~Zhang, and L.~Zhang, ``Image super-resolution: The techniques, applications, and future,'' \emph{Signal Processing}, vol. 128, pp. 389--408, 2016.

\bibitem{ZhangIsola2018}
R.~Zhang, P.~Isola, A.~A. Efros, E.~Shechtman, and O.~Wang, ``The unreasonable effectiveness of deep features as a perceptual metric,'' in \emph{Proc. IEEE/CVF CVPR}, 2018.

\end{thebibliography}
\end{document}